\documentclass[10pt,twocolumn,letterpaper]{article}

\usepackage{cvpr}
\usepackage{times}
\usepackage{epsfig}
\usepackage{graphicx}
\usepackage{amsmath}
\usepackage{amssymb}

\usepackage{mathptmx}

\usepackage{xspace}
\usepackage{color}
\usepackage{colortbl}

\usepackage{booktabs}                   

\definecolor{gray}{rgb}{0.35,0.35,0.35}
\definecolor{yellow}{rgb}{1,1,0.25}
\definecolor{MyBlue}{rgb}{0,0.2,0.8}
\definecolor{MyRed}{rgb}{0.8,0.2,0}
\definecolor{Red}{rgb}{1,0,0}
\definecolor{LightCyan}{rgb}{0.88,1,1}
\definecolor{MyGreen}{rgb}{0.0,0.5,0.1}
\definecolor{LightGray}{rgb}{0.9,0.9,0.9}

\def\red#1{\textcolor{red}{#1}}
\def\blue#1{\textcolor{blue}{#1}}

\newlength\paramargin
\newlength\figmargin
\newlength\secmargin

\usepackage{graphicx}
\graphicspath{{figures/}}

\usepackage{array}
\newcolumntype{L}[1]{>{\raggedright\let\newline\\\arraybackslash\hspace{0pt}}m{#1}}
\newcolumntype{C}[1]{>{\centering\let\newline\\\arraybackslash\hspace{0pt}}m{#1}}
\newcolumntype{R}[1]{>{\raggedleft\let\newline\\\arraybackslash\hspace{0pt}}m{#1}}

\long\def\ignorethis#1{}
\def\@onedot{\ifx\@let@token.\else.\null\fi\xspace}
\def\ie{i.e.}
\def\eg{e.g.}
\def\etal{et~al.\xspace}

\newcommand{\figref}[1]{Figure~\ref{fig:#1}}
\newcommand{\tabref}[1]{Table~\ref{tab:#1}} 

\renewcommand{\paragraph}[1]{\noindent\textbf{#1}}
\usepackage{subfigure}
\usepackage{multirow}
\usepackage{adjustbox}
\usepackage{caption}
\usepackage{enumitem}
\usepackage{float}

\usepackage[pagebackref=true,breaklinks=true,letterpaper=true,colorlinks,citecolor=blue,linkcolor=blue,bookmarks=false]{hyperref}

\cvprfinalcopy 

\def\cvprPaperID{192} 

\begin{document}
	\title{Deep Laplacian Pyramid Networks for Fast and Accurate Super-Resolution}
	
	\author{
		Wei-Sheng Lai$^{1}$ 
		\hspace{25pt} 
		Jia-Bin Huang$^{2}$ 
		\hspace{25pt} 
		Narendra Ahuja$^{3}$
		\hspace{25pt} 
		Ming-Hsuan Yang$^{1}$
		\\
		$^1$University of California, Merced
		\hspace{20pt} 
		$^2$Virginia Tech
		\hspace{20pt} 
		$^3$University of Illinois, Urbana-Champaign \\
		{\small
			\url{http://vllab.ucmerced.edu/wlai24/LapSRN}
		}
	}
	\maketitle
	\begin{abstract}
		Convolutional neural networks have recently demonstrated high-quality reconstruction for single-image super-resolution.
		In this paper, we propose the Laplacian Pyramid Super-Resolution Network (LapSRN) to progressively reconstruct the sub-band residuals of high-resolution images.
		At each pyramid level, our model takes coarse-resolution feature maps as input, predicts the high-frequency residuals, and uses transposed convolutions for upsampling to the finer level.
		Our method does not require the bicubic interpolation as the pre-processing step and thus dramatically reduces the computational complexity.
		We train the proposed LapSRN with deep supervision using a robust Charbonnier loss function and achieve high-quality reconstruction.
		Furthermore, our network generates multi-scale predictions in one feed-forward pass through the progressive reconstruction, thereby facilitates resource-aware applications.
		Extensive quantitative and qualitative evaluations on benchmark datasets show that the proposed algorithm performs favorably against the state-of-the-art methods in terms of speed and accuracy.
	\end{abstract}
	
	
	\section{Introduction}
	Single-image super-resolution (SR) aims to reconstruct a high-resolution (HR) image from a single low-resolution (LR) input image.
	In recent years, example-based SR methods have demonstrated the state-of-the-art performance by learning a mapping from LR to HR image patches using large image databases.
	Numerous learning algorithms have been applied to learn such a mapping, 
	including dictionary learning~\cite{Yang-CVPR-2008,Yang-TIP-2010}, local linear regression~\cite{A+,Yang-ICCV-2013}, and random forest~\cite{RFL}.

	Recently, Dong \etal~\cite{SRCNN} propose a Super-Resolution Convolutional Neural Network (SRCNN) to learn a nonlinear LR-to-HR mapping.
	The network is extended to embed a sparse coding-based network~\cite{SCN} or use a deeper structure~\cite{VDSR}. 
	While these models demonstrate promising results, there are three main issues.
	First, existing methods use a pre-defined upsampling operator, \eg, bicubic interpolation, to upscale input images to the desired spatial resolution before applying the network for prediction.
	This pre-processing step increases unnecessary computational cost and often results in visible reconstruction artifacts.
	Several algorithms accelerate SRCNN by performing convolution on LR images and replacing the pre-defined upsampling operator with sub-pixel convolution~\cite{ESPCN} or transposed convolution~\cite{FSRCNN} (also named as deconvolution in some of the literature).
	These methods, however, use relatively small networks and cannot learn complicated mappings well due to the limited network capacity.
	Second, existing methods optimize the networks with an $\ell_2$ loss and thus inevitably generate blurry predictions.
	Since the $\ell_2$ loss fails to capture the underlying multi-modal distributions of HR patches (\ie, the same LR patch may have many corresponding HR patches), the reconstructed HR images are often overly-smooth and not close to human visual perception on natural images. 
	Third, most methods reconstruct HR images \textit{in one upsampling step}, which increases the difficulties of training for large scaling factors (e.g., $8\times$).
	In addition, existing methods cannot generate intermediate SR predictions at multiple resolutions.
	As a result, one needs to train a large variety of models for various applications with different desired upsampling scales and computational loads.
	
	\begin{figure*}
		\centering
		\footnotesize
		\begin{tabular}{ccc}
			\centering
			\begin{adjustbox}{valign=b}
				\begin{tabular}{c}
					\includegraphics[height=1.7cm]{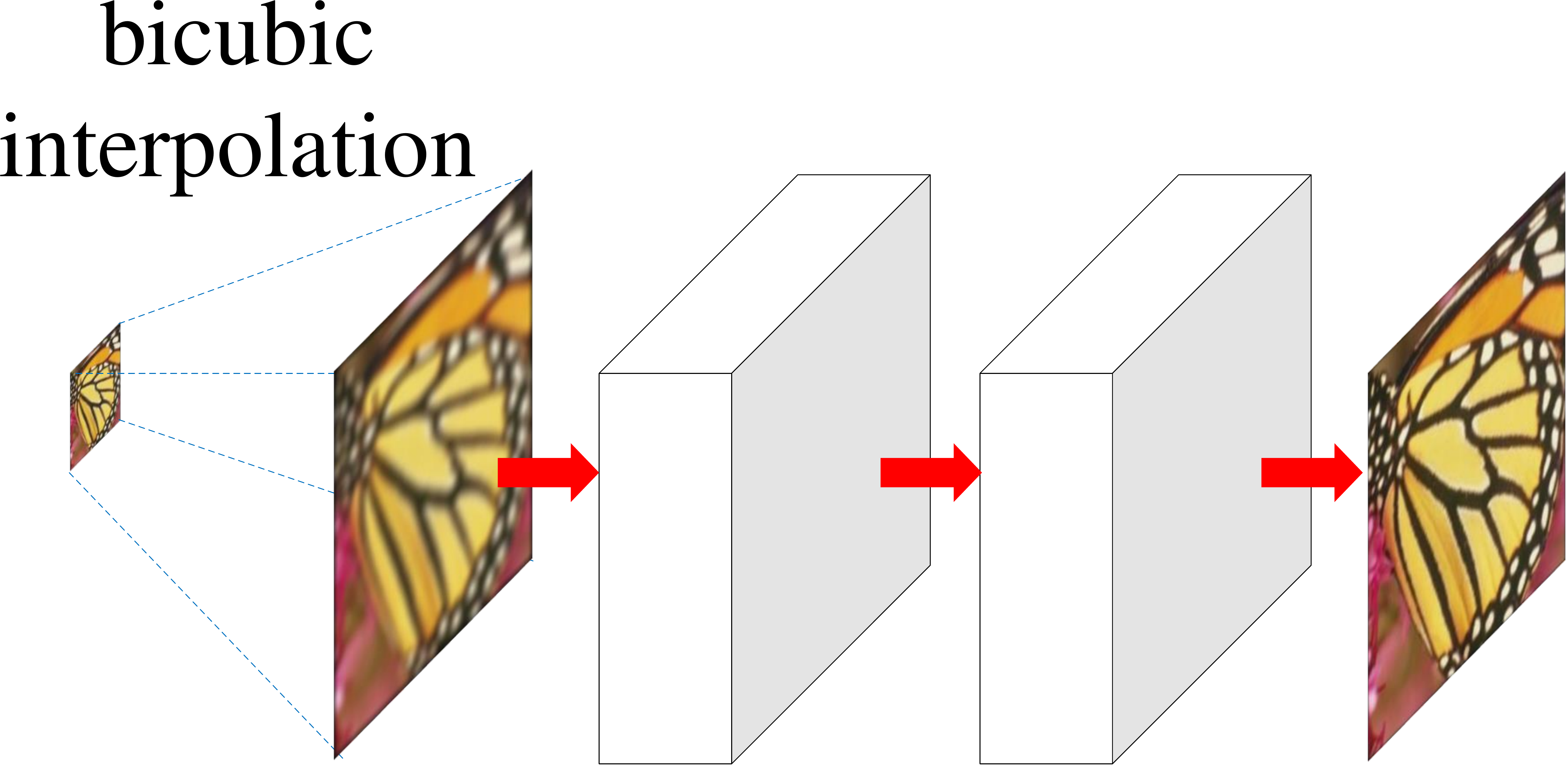}
					\\[1ex]
					(a) SRCNN~\cite{SRCNN}
					\\[3ex]
					\includegraphics[height=1.4cm]{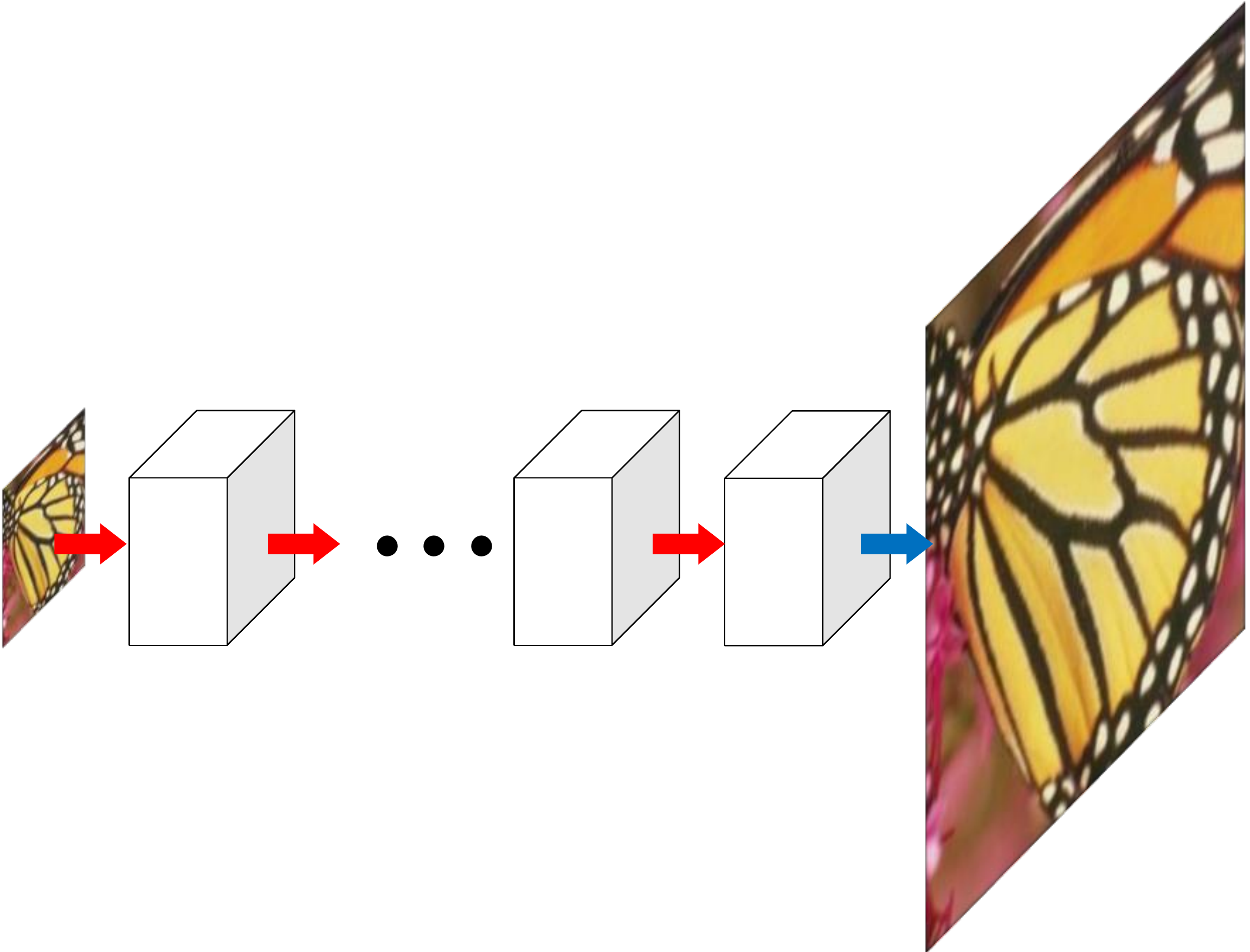}
					\\[1ex]
					(b) FSRCNN~\cite{FSRCNN}
				\end{tabular}
			\end{adjustbox}
			& 
			\begin{adjustbox}{valign=b}
				\begin{tabular}{c}
					\includegraphics[height=1.7cm]{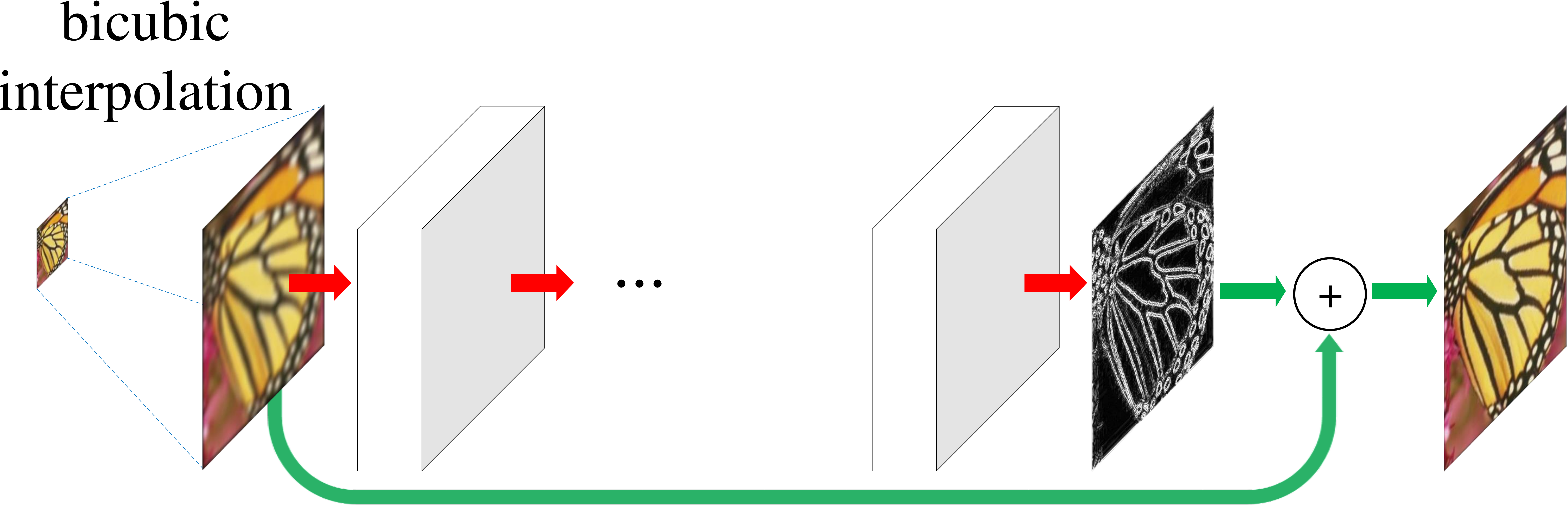}
					\\[1ex]
					(c) VDSR~\cite{VDSR}
					\\[1ex]
					\includegraphics[height=1.7cm]{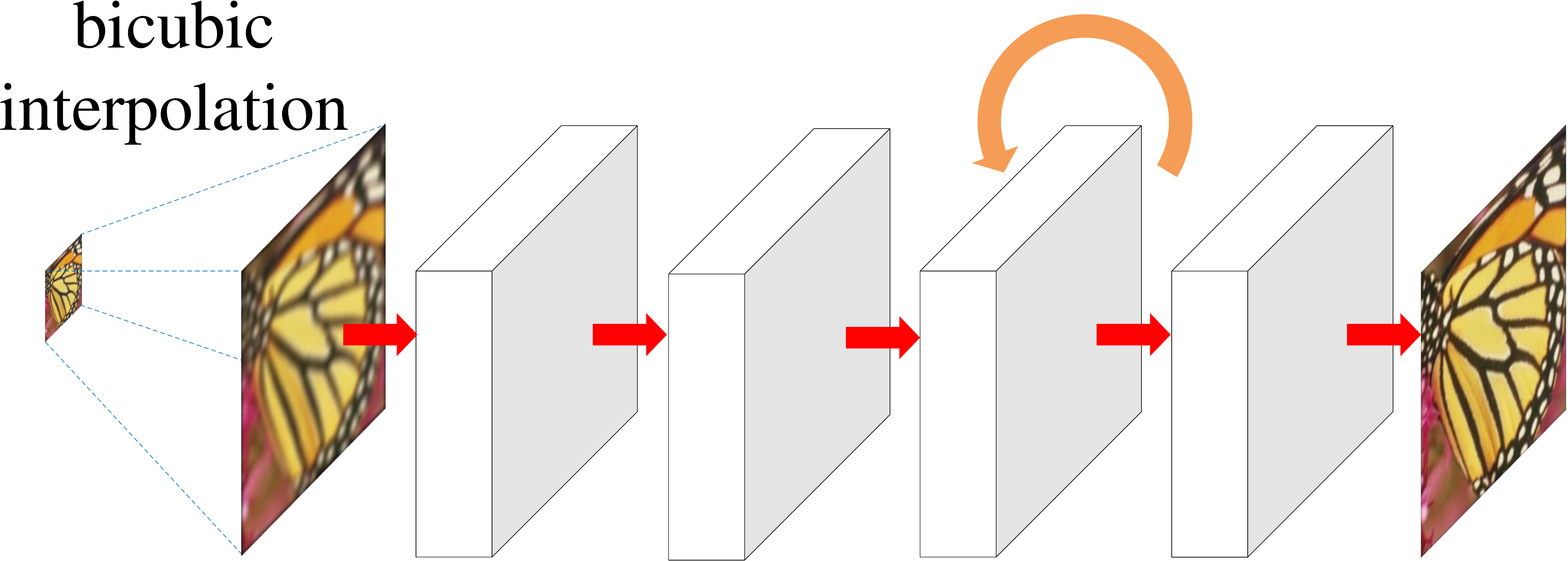}
					\\[1ex]
					(d) DRCN~\cite{DRCN}
				\end{tabular}
			\end{adjustbox}
			&
			\begin{adjustbox}{valign=b}
				\begin{tabular}{c}
					\includegraphics[height=4.5cm]{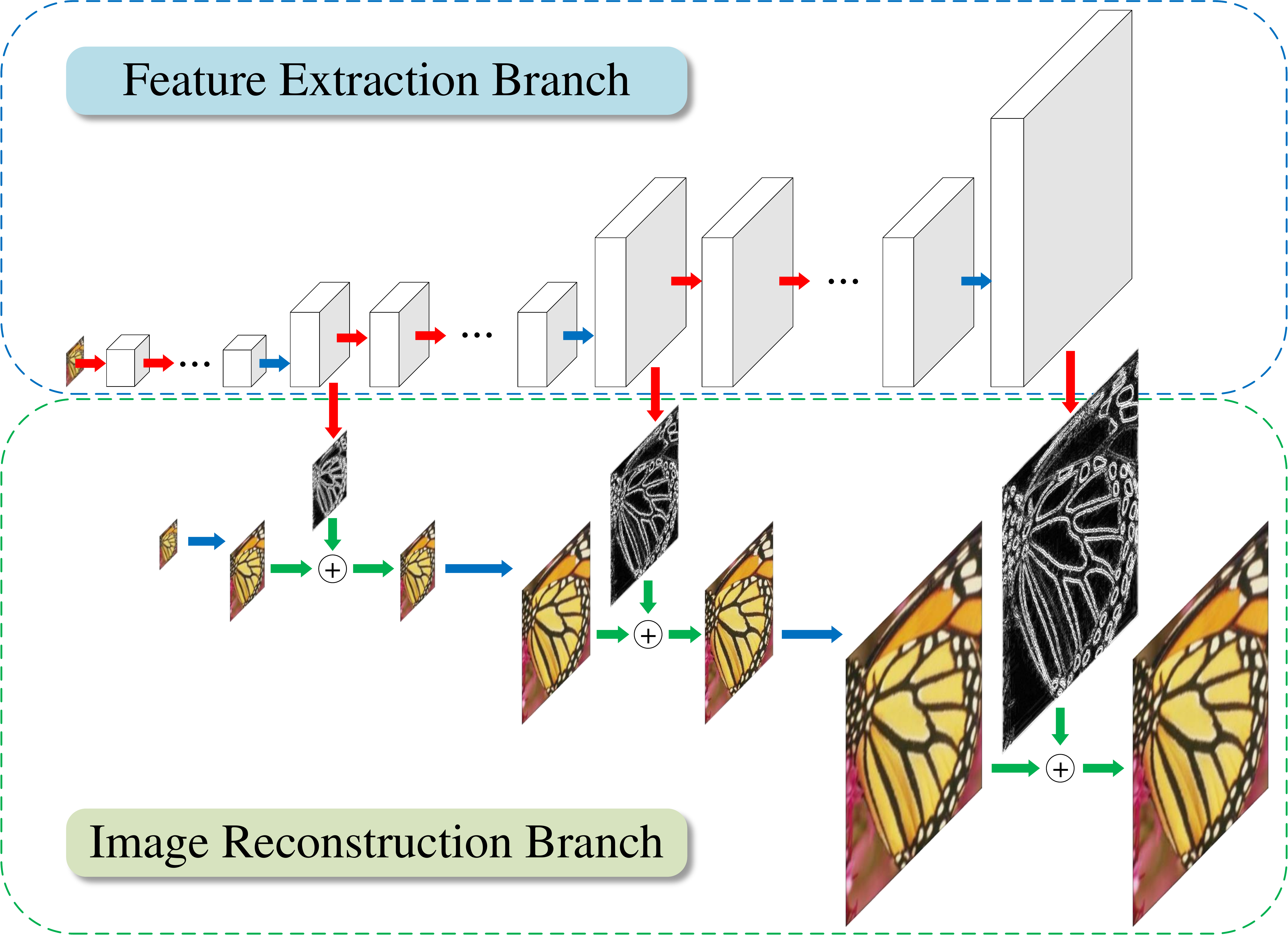}
					\\[1ex]
					(e) LapSRN (ours)
				\end{tabular}
			\end{adjustbox}
		\end{tabular}
		\vspace{-0.2cm}
		\caption{
			Network architectures of SRCNN~\cite{SRCNN}, FSRCNN~\cite{FSRCNN}, VDSR~\cite{VDSR}, DRCN~\cite{DRCN} and the proposed LapSRN. Red arrows indicate convolutional layers. Blue arrows indicate transposed convolutions (upsampling). Green arrows denote element-wise addition operators, and the orange arrow indicates recurrent layers.}
		\label{fig:CNN-compare}
		\vspace{-0.2cm}
	\end{figure*}
	
	To address these drawbacks, we propose the Laplacian Pyramid Super-Resolution Network (LapSRN) based on a cascade of convolutional neural networks (CNNs).
	Our network takes an LR image as input and progressively predicts the sub-band residuals in a coarse-to-fine fashion. 
	At each level, we first apply a cascade of convolutional layers to extract feature maps.
	We then use a transposed convolutional layer for upsampling the feature maps to a finer level.
	Finally, we use a convolutional layer to predict the sub-band residuals (the differences between the upsampled image and the ground truth HR image at the respective level).
	The predicted residuals at each level are used to efficiently reconstruct the HR image through upsampling and addition operations.
	While the proposed LapSRN consists of a set of cascaded sub-networks, we train the network with a robust Charbonnier loss function in an end-to-end fashion (\ie, without stage-wise optimization).
	As depicted in~\figref{CNN-compare}(e), our network architecture naturally accommodates deep supervision (\ie, supervisory signals can be applied simultaneously at each level of the pyramid).
	%

	Our algorithm differs from existing CNN-based methods in the following three aspects: 
	
	\noindent (1) \textbf{Accuracy}. 
	The proposed LapSRN extracts feature maps directly from LR images and jointly optimizes the upsampling filters with deep convolutional layers to predict sub-band \emph{residuals}.
	The deep supervision with the Charbonnier loss improves the performance thanks to the ability to better handle outliers.
	As a result, our model has a large capacity to learn complicated mappings and effectively reduces the undesired visual artifacts.
	
	\noindent (2) \textbf{Speed}.
	Our LapSRN embraces both fast processing speed and high capacity of deep networks.
	Experimental results demonstrate that our method is faster than several CNN based super-resolution models, e.g., SRCNN~\cite{SRCNN}, SCN~\cite{SCN}, VDSR~\cite{VDSR}, and DRCN~\cite{DRCN}.
	Similar to FSRCNN~\cite{FSRCNN}, our LapSRN achieves real-time speed on most of the evaluated datasets.
	In addition, our method provides significantly better reconstruction accuracy.
	
	\noindent (3) \textbf{Progressive reconstruction}. 
	Our model generates multiple intermediate SR predictions in \emph{one} feed-forward pass through progressive reconstruction using the Laplacian pyramid. 
	This characteristic renders our technique applicable to a wide range of applications that require resource-aware adaptability. 
	For example, the same network can be used to enhance the spatial resolution of videos depending on the available computational resources.
	For scenarios with limited computing resources, 
	our $8\times$ model can still perform 2$\times$ or $4\times$ SR by simply bypassing the computation of residuals at finer levels. 
	Existing CNN-based methods, however, do not offer such flexibility.

	\begin{table*}
		\centering
		\caption{
			Comparisons of CNN based SR algorithms: SRCNN~\cite{SRCNN}, FSRCNN~\cite{FSRCNN}, SCN~\cite{SCN}, ESPCN~\cite{ESPCN}, VDSR~\cite{VDSR}, and the proposed LapSRN.
			The number of layers includes both convolution and transposed convolution.
			%
			%
			Methods with direct reconstruction performs one-step upsampling (with bicubic interpolation or transposed convolution) from LR to HR images, while progressive reconstruction predicts HR images in multiple steps.
		}
		\vspace{-0.1cm}
		\begin{tabular}{llcccc}
			\toprule
			Method & 
			Network input & 
			$\#$Layers & 
			Residual learning&
			Reconstruction &
			Loss function\\
			\midrule
			SRCNN~\cite{SRCNN} 
			& LR + bicubic & 3 & No & Direct & L2 \\
			FSRCNN~\cite{FSRCNN} 
			& LR & 8 & No & Direct & L2 \\
			SCN~\cite{SCN} 
			& LR + bicubic & 5 & No & Progressive &L2 \\
			ESPCN~\cite{ESPCN} 
			& LR & 3 & No & Direct & L2 \\
			VDSR~\cite{VDSR} 
			& LR + bicubic & 20 & Yes & Direct & L2 \\
			DRCN~\cite{DRCN} 
			& LR + bicubic & 5 (recursive) & No & Direct & L2 \\
			LapSRN (ours) 
			& LR & 27 & Yes &Progressive& Charbonnier \\
			\bottomrule
		\end{tabular}
		\label{tab:CNN-compare}
		\vspace{-0.3cm}
	\end{table*}

	\section{Related Work and Problem Context}
	
	Numerous single-image super-resolution methods have been proposed in the literature.
	Here we focus our discussion on recent example-based approaches.
	
	\noindent\paragraph{SR based on internal databases.}
	Several methods~\cite{Freedman-TOG-2011,Glasner-ICCV-2009} exploit the self-similarity property in natural images and construct LR-HR patch pairs based on the scale-space pyramid of the low-resolution input image.
	While internal databases contain more relevant training patches than external image databases, 
	the number of LR-HR patch pairs may not be sufficient to cover large textural variations in an image.
	Singh~\etal~\cite{Singh-ACCV-2014} decompose patches into directional frequency sub-bands and determine better matches in each sub-band pyramid independently.
	Huang~\etal~\cite{Huang-CVPR-2015} extend the patch search space to accommodate the affine transform and perspective deformation.
	The main drawback of SR methods based on internal databases is that they are typically slow due to the heavy computational cost of patch search in the scale-space pyramid.
	
	\paragraph{SR based on external databases.}
	Numerous SR methods learn the LR-HR mapping with image pairs collected from external databases using supervised learning algorithms, such as nearest neighbor~\cite{Freeman-CGA-2002}, manifold embedding~\cite{Bevilacqua-BMVC-2012,Chang-CVPR-2004}, kernel ridge regression~\cite{Kim-PAMI-2010}, and sparse representation~\cite{Yang-CVPR-2008,Yang-TIP-2010,Zeyde-2010}.
	Instead of directly modeling the complex patch space over the entire database, several methods partition the image database by K-means~\cite{Yang-ICCV-2013}, sparse dictionary~\cite{A+} or random forest~\cite{RFL}, and learn locally linear regressors for each cluster.

	\noindent\paragraph{Convolutional neural networks based SR.} 
	In contrast to modeling the LR-HR mapping in the \emph{patch} space, SRCNN~\cite{SRCNN} jointly optimize all the steps and learn the non-linear mapping in the \emph{image} space.
	The VDSR network~\cite{VDSR} demonstrates significant improvement over SRCNN~\cite{SRCNN} by increasing the network depth from 3 to 20 convolutional layers.
	To facilitate training a deeper model with a fast convergence speed, VDSR trains the network to predict the residuals rather the actual pixel values.
	Wang \etal~\cite{SCN} combine the domain knowledge of sparse coding with a deep CNN and train a cascade network (SCN) to upsample images to the desired scale factor progressively.
	Kim~\etal~\cite{DRCN} propose a shallow network with deeply recursive layers (DRCN) to reduce the number of parameters.

	To achieve real-time performance, the ESPCN network~\cite{ESPCN} extracts feature maps in the LR space and replaces the bicubic upsampling operation with an efficient sub-pixel convolution.
	The FSRCNN network~\cite{FSRCNN} adopts a similar idea and uses a hourglass-shaped CNN with more layers but fewer parameters than that in ESPCN.
	All the above CNN-based SR methods optimize networks with an $\ell_2$ loss function, which often leads to overly-smooth results that do not correlate well with human perception.
	In the context of SR, we demonstrate that the $\ell_2$ loss is less effective for learning and predicting sparse residuals.

	We compare the network structures of SRCNN, FSRCNN, VDSR, DRCN and our LapSRN in~\figref{CNN-compare} and list the main differences among existing CNN-based methods and the proposed framework in~\tabref{CNN-compare}.
	Our approach builds upon existing CNN-based SR algorithms with three main differences.
	First, we jointly learn residuals and upsampling filters with convolutional and transposed convolutional layers.
	Using the learned upsampling filters not only effectively suppresses reconstruction artifacts caused by the bicubic interpolation, but also dramatically reduces the computational complexity.
	Second, we optimize the deep network using a robust Charbonnier loss function instead of the $\ell_2$ loss to handle outliers and improve the reconstruction accuracy.
	Third, as the proposed LapSRN progressively reconstructs HR images, the same model can be used for applications that require different scale factors by truncating the network up to a certain level.

	\noindent\paragraph{Laplacian pyramid.} 
	The Laplacian pyramid has been used in a wide range of applications, such as image blending~\cite{Burt-1983}, texture synthesis~\cite{Heeger-TOG-1995}, edge-aware filtering~\cite{Paris-TOG-2011} and semantic segmentation~\cite{Ghiasi-ECCV-2016,Pinheiro-ECCV-2016}.
	Denton~\etal propose a generative model based on a Laplacian pyramid framework (LAPGAN) to generate realistic images in~\cite{LAPGAN}, which is the most related to our work.
	However, the proposed LapSRN differs from LAPGAN in three aspects.
	
	First, LAPGAN is a \emph{generative model} which is designed to synthesize diverse natural images from random noise and sample inputs.
	On the contrary, our LapSRN is a super-resolution model that predicts a particular HR image based on the given LR image.
	LAPGAN uses a cross-entropy loss function to encourage the output images to respect the data distribution of training datasets. 
	In contrast, we use the Charbonnier penalty function to penalize the deviation of the prediction from the ground truth sub-band residuals.
	%
	
	Second, the sub-networks of LAPGAN are \emph{independent} (i.e., no weight sharing).
	As a result, the network capacity is limited by the depth of each sub-network.
	Unlike LAPGAN, the convolutional layers at each level in LapSRN are \emph{connected} through multi-channel transposed convolutional layers.
	The residual images at a higher level are therefore predicted by a deeper network with shared feature representations at lower levels.
	The feature sharing at lower levels increases the non-linearity at finer convolutional layers to learn complex mappings.
	Also, the sub-networks in LAPGAN are \emph{independently trained}.
	On the other hand, all the convolutional filters for feature extraction, upsampling, and residual prediction layers in the LapSRN are \emph{jointly trained} in an end-to-end, deeply supervised fashion.
	%
	
	Third, LAPGAN applies convolutions on the upsampled images, so the speed depends on the size of HR images.
	On the contrary, our design of LapSRN effectively increases the size of the receptive field and accelerates the speed by extracting features from the LR space.
	We provide comparisons with LAPGAN in the supplementary material.

	\noindent\paragraph{Adversarial training.} 
	The SRGAN method~\cite{SRGAN} optimizes the network using the perceptual loss~\cite{Johnson-ECCV-2016} and the adversarial loss for photo-realistic SR.
	We note that our LapSRN can be easily extended to the adversarial training framework.
	As it is not our contribution, we provide experiments on the adversarial loss in the supplementary material.

	\clearpage
	\section{Deep Laplacian Pyramid Network for SR}
	\label{sec:framework}
	\vspace{-0.1cm}
	In this section, we describe the design methodology of the proposed Laplacian pyramid network, the optimization using robust loss functions with deep supervision, and the details for network training.
	
	\subsection{Network architecture}
	\label{sec:architecture}
	\vspace{-0.1cm}
	We propose to construct our network based on the Laplacian pyramid framework, as shown in \figref{CNN-compare}(e).
	Our model takes an LR image as input (rather than an upscaled version of the LR image) and progressively predicts residual images at $\log_2 S$ levels where $S$ is the scale factor.
	For example, the network consists of $3$ sub-networks for super-resolving an LR image at a scale factor of $8$.
	Our model has two branches: (1) feature extraction and (2) image reconstruction.
	%
	
	\noindent\textbf{Feature extraction.} 
	At level $s$, the feature extraction branch consists of $d$ convolutional layers and one transposed convolutional layer to upsample the extracted features by a scale of 2.
	The output of each transposed convolutional layer is connected to two different layers:
	(1) a convolutional layer for reconstructing a residual image at level $s$, 
	and (2) a convolutional layer for extracting features at the finer level $s+1$.
	Note that we perform the feature extraction at the \emph{coarse} resolution and generate feature maps at the \emph{finer} resolution with only one transposed convolutional layer.
	In contrast to existing networks that perform all feature extraction and reconstruction at the fine resolution, our network design significantly reduces the computational complexity.
	Note that the feature representations at lower levels are shared with higher levels, and thus can increase the non-linearity of the network to learn complex mappings at the finer levels.
	
	\noindent\textbf{Image reconstruction.} 
	At level $s$, the input image is upsampled by a scale of 2 with a transposed convolutional (upsampling) layer.
	We initialize this layer with the bilinear kernel and allow it to be jointly optimized with all the other layers.
	The upsampled image is then combined (using element-wise summation) with the predicted residual image from the feature extraction branch to produce a high-resolution output image.
	The output HR image at level $s$ is then fed into the image reconstruction branch of level $s+1$.
	The entire network is a cascade of CNNs with a similar structure at each level.

	\subsection{Loss function}
	Let $x$ be the input LR image and $\theta$ be the set of network parameters to be optimized.
	Our goal is to learn a mapping function $f$ for generating a high-resolution image $\hat{y} = f(x; \theta)$ that is close to the ground truth HR image $y$.
	We denote the residual image at level $s$ by $r_s$, the upscaled LR image by $x_s$ and the corresponding HR images by $y_s$.
	The desired output HR images at level $s$ is modeled by $y_s = x_s + r_s$.
	We use the bicubic downsampling to resize the ground truth HR image $y$ to $y_s$ at each level.
	Instead of minimizing the mean square errors between $y_s$ and $\hat{y}_s$, we propose to use a robust loss function to handle outliers.
	The overall loss function is defined as:
	\begin{align}
	\mathcal{L}(\hat{y}, y; \theta) 
	=&~ \frac{1}{N} \sum_{i=1}^N\sum_{s=1}^L \rho\left( \hat{y}_s^{(i)} - y_s^{(i)} \right) \nonumber\\
	=&~ \frac{1}{N} \sum_{i=1}^N\sum_{s=1}^L \rho\left( (\hat{y}_s^{(i)} - x_s^{(i)}) - r_s^{(i)} \right),
	\end{align}
	where $\rho(x) = \sqrt{x^2 + \epsilon^2}$ is the Charbonnier penalty function (a differentiable variant of $\ell_1$ norm)~\cite{Bruhn-IJCV-2005}, 
	$N$ is the number of training samples in each batch,
	and $L$ is the number of level in our pyramid.
	We empirically set $\epsilon$ to $1e-3$.

	In the proposed LapSRN, each level $s$ has its loss function and the corresponding ground truth HR image $y_s$.
	This multi-loss structure resembles the deeply-supervised nets for classification~\cite{Lee-AISTATS-2015} and edge detection~\cite{Xie-CVPR-2015}.
	However, the labels used to supervise intermediate layers in~\cite{Lee-AISTATS-2015,Xie-CVPR-2015} are the \emph{same} across the networks.
	In our model, we use \emph{different} scales of HR images at the corresponding level as supervision.
	The deep supervision guides the network training to predict sub-band residual images at different levels and produce multi-scale output images.
	For example, 
	our $8\times$ model can produce $2\times$, $4\times$ and $8\times$ super-resolution results in \emph{one} feed-forward pass.
	This property is particularly useful for resource-aware applications, e.g., mobile devices or network applications.
	
	\subsection{Implementation and training details}
	\vspace{-0.1cm}
	In the proposed LapSRN, each convolutional layer consists of 64 filters with the size of $3\times3$. 
	We initialize the convolutional filters using the method of He~\etal~\cite{He-ICCV-2015}. 
	The size of the transposed convolutional filters is $4 \times 4$ and the weights are initialized from a bilinear filter.
	All the convolutional and transposed convolutional layers (except the reconstruction layers) are followed by leaky rectified linear units (LReLUs) with a negative slope of 0.2.
	We pad zeros around the boundaries before applying convolution to keep the size of all feature maps the same as the input of each level.
	The convolutional filters have small spatial supports ($3\times3$). 
	However, we can achieve high non-linearity and increase the size of receptive fields with a deep structure.

	We use 91 images from Yang~\etal~\cite{Yang-TIP-2010} and 200 images from the training set of Berkeley Segmentation Dataset~\cite{BSDS} as our training data.
	The same training dataset is used in~\cite{VDSR,RFL} as well.
	In each training batch, we randomly sample $64$ patches with the size of $128 \times 128$.
	An epoch has $1,000$ iterations of back-propagation. 
	We augment the training data in three ways: 
	(1) \emph{Scaling}: randomly downscale between $[0.5, 1.0]$.
	(2) \emph{Rotation}: randomly rotate image by $90^\circ$, $180^\circ$, or $270^\circ$.
	(3) \emph{Flipping}: flip images horizontally or vertically with a probability of $0.5$. 
	Following the protocol of existing methods~\cite{SRCNN,VDSR}, we generate the LR training patches using the bicubic downsampling.
	We train our model with the MatConvNet toolbox~\cite{Vedaldi-ACMMM-2015}.
	We set momentum parameter to $0.9$ and the weight decay to $1e-4$. 
	The learning rate is initialized to $1e-5$ for all layers and decreased by a factor of 2 for every 50 epochs.
	%
	
	\section{Experiment Results}
	\label{sec:experiments}
	We first analyze the contributions of different components of the proposed network.
	We then compare our LapSRN with state-of-the-art algorithms on five benchmark datasets and demonstrate the applications of our method on super-resolving real-world photos and videos.
	
	\subsection{Model analysis} 

\begin{figure}[t]
		\centering
		\includegraphics[width=0.75\linewidth]{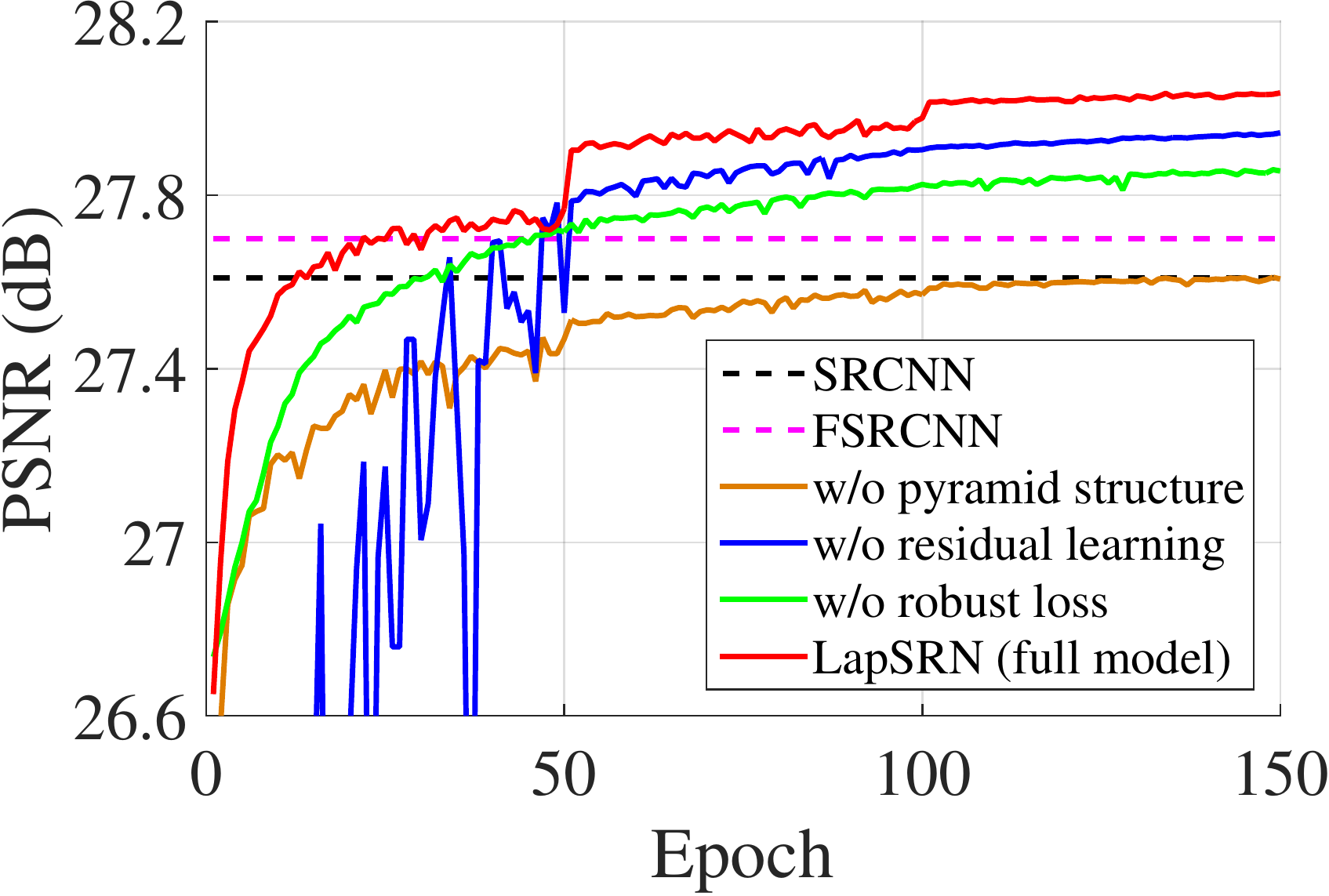} 
		\vspace{-0.3cm}
		\caption{
			Convergence analysis on the pyramid structure, loss functions and residual learning.
			Our LapSRN converges faster and achieves improved performance.
		}
		\label{fig:compare_loss_curve}
	\end{figure}

	\noindent\textbf{Residual learning.} 
	To demonstrate the effect of residual learning, we remove the image reconstruction branch and directly predict the HR images at each level.
	\figref{compare_loss_curve} shows the convergence curves in terms of PSNR on the \textsc{Set14} for $4\times$ SR.
	The performance of the ``non-residual'' network (blue curve) converges slowly and fluctuates significantly. 
	The proposed LapSRN (red curve), on the other hand, outperforms SRCNN within 10 epochs.
	
	\begin{table}[t]
		\centering
		\caption{
			Ablation study of pyramid structures, loss functions, and residual learning.
			We replace each component with the one used in existing methods, and observe performance (PSNR) drop on both \textsc{Set5} and \textsc{Set14}.
		}
		\vspace{-2mm}
		\begin{tabular}{ccc|cc}
			\toprule
			Residual & Pyramid & Loss & \textsc{Set5} & \textsc{Set14} \\
			\midrule
			\checkmark & & Robust & 30.58 & 27.61 \\
			& \checkmark & Robust & 31.10 & 27.94 \\
			\checkmark & \checkmark & $\ell_2$ & 30.93 & 27.86 \\
			\checkmark & \checkmark & Robust & \textbf{31.28} & \textbf{28.04} \\
			\bottomrule
		\end{tabular}
		\label{tab:component-effect}
		\vspace{-5mm}
	\end{table}

	\begin{figure}[t]
		\scriptsize
		\centering
		\begin{tabular}{cc}
			\hspace{-0.45cm}
			\begin{adjustbox}{valign=t}
				\begin{tabular}{c}
					\includegraphics[width=0.29\columnwidth]{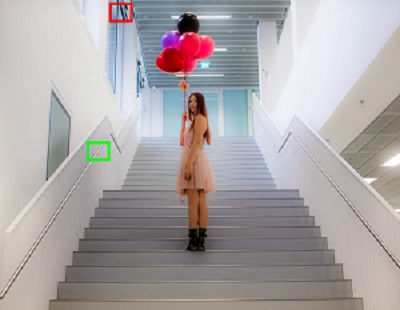}
					\\
					(a)
				\end{tabular}
			\end{adjustbox}
			\hspace{-0.45cm}
			\begin{adjustbox}{valign=t}
				\begin{tabular}{ccccc}
					\includegraphics[width=0.13\columnwidth]{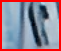}
					\hspace{-0.4cm} &
					\includegraphics[width=0.13\columnwidth]{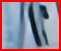}
					\hspace{-0.4cm} &
					\includegraphics[width=0.13\columnwidth]{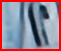}
					\hspace{-0.4cm} &
					\includegraphics[width=0.13\columnwidth]{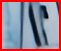} 
					\hspace{-0.4cm} &
					\includegraphics[width=0.13\columnwidth]{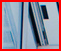} 
					\\
					\includegraphics[width=0.13\columnwidth]{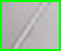}
					\hspace{-0.4cm} &
					\includegraphics[width=0.13\columnwidth]{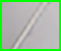}
					\hspace{-0.4cm} &
					\includegraphics[width=0.13\columnwidth]{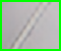}
					\hspace{-0.4cm} &
					\includegraphics[width=0.13\columnwidth]{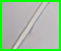} 
					\hspace{-0.4cm} &
					\includegraphics[width=0.13\columnwidth]{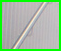} 
					\\
					(b) \hspace{-0.4cm} & 
					(c) \hspace{-0.4cm} & 
					(d) \hspace{-0.4cm} & 
					(e) \hspace{-0.4cm} & 
					(f) 
					\\
				\end{tabular}
			\end{adjustbox}
		\end{tabular}
		\vspace{-2mm}
		\caption{Contribution of different components in the proposed network. 
			(a) HR image. (b) w/o pyramid structure (c) w/o residual learning (d) w/o robust loss (e) full model (f) ground truth.
		}
		\label{fig:compare_loss_visual}
		\vspace{-0.2cm}
	\end{figure}
	
	\begin{table}[t]
		\centering
		\caption{
			Trade-off between performance and speed on the depth at each level of the proposed network.
		}
		\vspace{-2mm}
		\begin{tabular}{c|cccc}
			\toprule
			\multirow{2}{*}{Depth} & 
			\multicolumn{2}{c}{\textsc{Set5}} & 
			\multicolumn{2}{c}{\textsc{Set14}} \\ 
			& PSNR & Second & PSNR & Second \\ \midrule
			3 & 31.15 & 0.036 & 27.98 & 0.036 \\
			5 & 31.28 & 0.044 & 28.04 & 0.042 \\
			10 & 31.37 & 0.050 & 28.11 & 0.051 \\
			15 & 31.45 & 0.077 & 28.16 & 0.071 \\
			\bottomrule
		\end{tabular}
		\label{tab:depth_speed}
		\vspace{-5mm}
	\end{table}

	\noindent\textbf{Loss function.} 
	To validate the effect of the Charbonnier loss function, we train the proposed network with the $\ell_2$ loss function.
	We use a larger learning rate ($1e-4$) since the gradient magnitude of the $\ell_2$ loss is smaller. 
	As illustrated in~\figref{compare_loss_curve}, the network optimized with $\ell_2$ loss (green curve) requires more iterations to achieve comparable performance with SRCNN. 
	In~\figref{compare_loss_visual}(d), we show that the network trained with the $\ell_2$ loss generates SR results with more ringing artifacts. 
	In contrast, the SR images reconstruct by the proposed algorithm (\figref{compare_loss_visual}(e)) contain relatively clean and sharp details. 

	\noindent\textbf{Pyramid structure.}
	By removing the pyramid structure, our model falls back to a network similar to FSRCNN but with the residual learning.
	To use the same number of convolutional layers as LapSRN, we train a network with 10 convolutional layers and one transposed convolutional layer.
	The quantitative results in~\tabref{component-effect} shows that the pyramid structure leads to moderate performance improvement (e.g. 0.7 dB on \textsc{Set5} and 0.4 dB on \textsc{Set14}).

	\noindent\textbf{Network depth.} 
	We train the proposed model with different depth, $d = 3, 5, 10, 15$, at each level and show the trade-offs between performance and speed in~\tabref{depth_speed}.
	In general, deep networks perform better shallow ones at the expense of increased computational cost.
	We choose $d = 10$ for our $2\times$ and $4\times$ SR models to strike a balance between performance and speed.
	We show that the speed of our LapSRN with $d = 10$ is faster than most of the existing CNN-based SR algorithms (see~\figref{time-chart}).
	For $8\times$ model, we choose $d = 5$ because we do not observe significant performance gain by using more convolutional layers.

	\begin{figure*}[t]
		\scriptsize
		\centering
		\begin{tabular}{cc}
			\begin{adjustbox}{valign=t}
				\begin{tabular}{c}
					\includegraphics[width=0.23\textwidth]{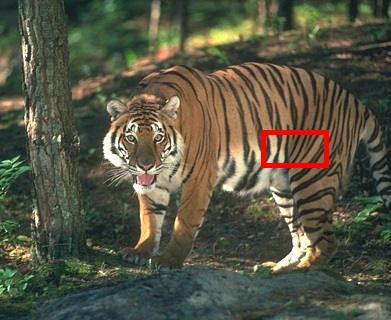}
					\\
					Ground-truth HR
				\end{tabular}
			\end{adjustbox}
			\hspace{-0.32cm}
			\begin{adjustbox}{valign=t}
				\begin{tabular}{cccc}
					\includegraphics[width=0.165\textwidth]{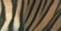} \hspace{-0.25cm} &
					\includegraphics[width=0.165\textwidth]{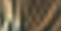} \hspace{-0.25cm} &
					\includegraphics[width=0.165\textwidth]{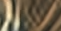} \hspace{-0.25cm} &
					\includegraphics[width=0.165\textwidth]{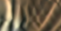}
					\\
					HR (PSNR, SSIM)\hspace{-0.25cm} &
					Bicubic (24.76, 0.6633)\hspace{-0.25cm} &
					A+~\cite{A+} (25.59, 0.7139)\hspace{-0.25cm} &
					SelfExSR~\cite{Huang-CVPR-2015} (25.45, 0.7087)
					\\
					\includegraphics[width=0.165\textwidth]{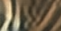} \hspace{-0.25cm} &
					\includegraphics[width=0.165\textwidth]{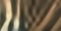} \hspace{-0.25cm} &
					\includegraphics[width=0.165\textwidth]{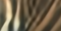} \hspace{-0.25cm} &
					\includegraphics[width=0.165\textwidth]{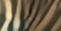} 
					\\ 
					FSRCNN~\cite{FSRCNN} (25.81, 0.7248)\hspace{-0.25cm} &
					VDSR~\cite{VDSR} (25.94, 0.7353)\hspace{-0.25cm} &
					DRCN~\cite{DRCN} (25.98, 0.7357)\hspace{-0.25cm} &
					Ours (\textbf{26.09}, \textbf{0.7403}) 
				\end{tabular}
			\end{adjustbox}
			\\ 
			\begin{adjustbox}{valign=t}
				\begin{tabular}{c}
					\includegraphics[width=0.23\textwidth]{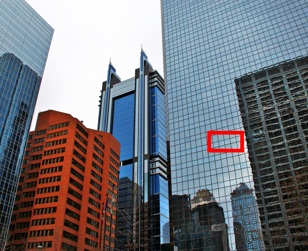}
					\\
					Ground-truth HR
				\end{tabular}
			\end{adjustbox}
			\hspace{-0.32cm}
			\begin{adjustbox}{valign=t}
				\begin{tabular}{cccc}
					\includegraphics[width=0.165\textwidth]{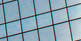} \hspace{-0.25cm} &
					\includegraphics[width=0.165\textwidth]{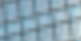} \hspace{-0.25cm} &
					\includegraphics[width=0.165\textwidth]{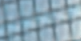} \hspace{-0.25cm} &
					\includegraphics[width=0.165\textwidth]{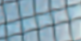}
					\\
					HR (PSNR, SSIM)\hspace{-0.25cm} &
					Bicubic (22.43, 0.5926)\hspace{-0.25cm} &
					A+~\cite{A+} (23.19, 0.6545)\hspace{-0.25cm} &
					SelfExSR~\cite{Huang-CVPR-2015} (23.88, 0.6961)
					\\
					\includegraphics[width=0.165\textwidth]{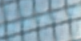} \hspace{-0.25cm} &
					\includegraphics[width=0.165\textwidth]{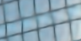} \hspace{-0.25cm} &
					\includegraphics[width=0.165\textwidth]{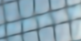} \hspace{-0.25cm} &
					\includegraphics[width=0.165\textwidth]{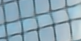} 
					\\
					FSRCNN~\cite{FSRCNN} (23.61, 0.6708)\hspace{-0.25cm} &
					VDSR~\cite{VDSR} (24.25, 0.7030)\hspace{-0.25cm} &
					DRCN~\cite{DRCN} (23.95, 0.6947)\hspace{-0.25cm} &
					Ours (\textbf{24.36}, \textbf{0.7200}) 
				\end{tabular}
			\end{adjustbox}
			\\
			\begin{adjustbox}{valign=t}
				\begin{tabular}{c}
					\includegraphics[width=0.23\textwidth]{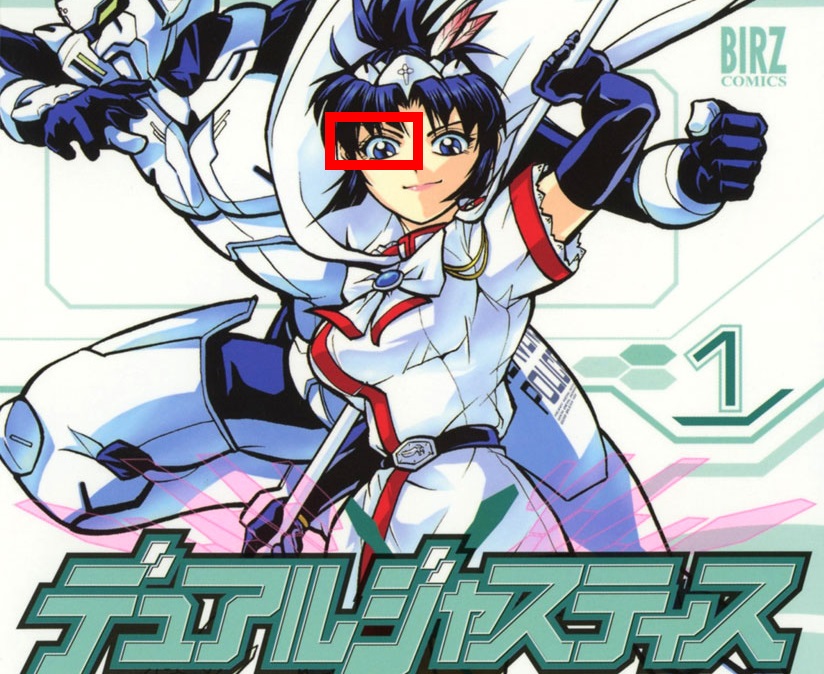}
					\\
					Ground-truth HR
				\end{tabular}
			\end{adjustbox}
			\hspace{-0.32cm}
			\begin{adjustbox}{valign=t}
				\begin{tabular}{cccc}
					\includegraphics[width=0.165\textwidth]{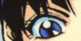} \hspace{-0.25cm} &
					\includegraphics[width=0.165\textwidth]{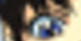} \hspace{-0.25cm} &
					\includegraphics[width=0.165\textwidth]{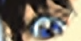} \hspace{-0.25cm} &
					\includegraphics[width=0.165\textwidth]{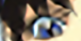}
					\\
					HR (PSNR, SSIM)\hspace{-0.25cm}&
					Bicubic (23.53, 0.8073)\hspace{-0.25cm}&
					A+~\cite{A+} (26.10, 0.8793)\hspace{-0.25cm}&
					SelfExSR~\cite{Huang-CVPR-2015} (26.75, 0.8960)
					\\
					\includegraphics[width=0.165\textwidth]{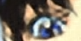} \hspace{-0.25cm} &
					\includegraphics[width=0.165\textwidth]{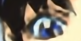} \hspace{-0.25cm} &
					\includegraphics[width=0.165\textwidth]{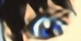} \hspace{-0.25cm} &
					\includegraphics[width=0.165\textwidth]{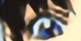} 
					\\
					FSRCNN~\cite{FSRCNN} (27.19, 0.8896) \hspace{-0.25cm} &
					VDSR~\cite{VDSR} (27.99, 0.9202) \hspace{-0.25cm} &
					DRCN~\cite{DRCN} (28.18, 0.9218) \hspace{-0.25cm} &
					Ours (\textbf{28.25}, \textbf{0.9224})
				\end{tabular}
			\end{adjustbox}
		\end{tabular}
		\vspace{-0.2cm}
		\caption{Visual comparison for $4\times$ SR on \textsc{BSDS100}, \textsc{Urban100} and \textsc{Manga109}.}
		\vspace{-0.1cm}
		\label{fig:result-4x}
	\end{figure*}

	\begin{figure*}
		\footnotesize
		\centering
		\begin{tabular}{cccc}
			\begin{adjustbox}{valign=t}
				\begin{tabular}{c}
					\includegraphics[width=0.4\columnwidth]{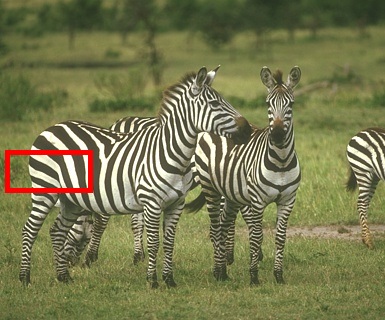}
					\\
					Ground-truth HR
				\end{tabular}
			\end{adjustbox}
			\hspace{-0.35cm}
			\begin{adjustbox}{valign=t}
				\begin{tabular}{cc}
					\includegraphics[width=0.265\columnwidth]{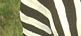} \hspace{-0.3cm} &
					\includegraphics[width=0.265\columnwidth]{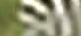}
					\\
					\hspace{0.2cm} HR&
					FSRCNN~\cite{FSRCNN}
					\\
					\hspace{0.2cm} (PSNR, SSIM)&
					(19.57, 0.5133) 
					\\
					\includegraphics[width=0.265\columnwidth]{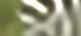} \hspace{-0.3cm} &
					\includegraphics[width=0.265\columnwidth]{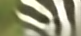} 
					\\
					\hspace{0.2cm} VDSR~\cite{VDSR}&
					LapSRN (ours)
					\\
					\hspace{0.2cm} (19.58, 0.5147)&
					(\textbf{19.75}, \textbf{0.5246}) 
				\end{tabular}
			\end{adjustbox}
			\begin{adjustbox}{valign=t}
				\begin{tabular}{c}
					\includegraphics[width=0.4\columnwidth]{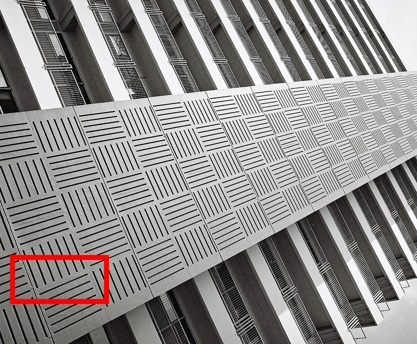}
					\\
					Ground-truth HR
				\end{tabular}
			\end{adjustbox}
			\hspace{-0.35cm}
			\begin{adjustbox}{valign=t}
				\begin{tabular}{cc}
					\includegraphics[width=0.265\columnwidth]{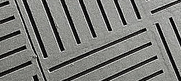} \hspace{-0.3cm} &
					\includegraphics[width=0.265\columnwidth]{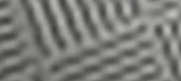}
					\\
					\hspace{0.2cm} HR &
					FSRCNN~\cite{FSRCNN}
					\\
					\hspace{0.2cm} (PSNR, SSIM)&
					(15.61, 0.3494) 
					\\
					\includegraphics[width=0.265\columnwidth]{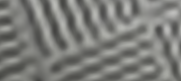} \hspace{-0.3cm} &
					\includegraphics[width=0.265\columnwidth]{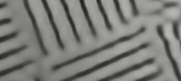} 
					\\
					\hspace{0.2cm} VDSR~\cite{VDSR}&
					LapSRN (ours)
					\\
					\hspace{0.2cm} (15.66, 0.3644)&
					(\textbf{15.72}, \textbf{0.3865}) 
				\end{tabular}
			\end{adjustbox}
		\end{tabular}
		\vspace{-2mm}
		\caption{
			Visual comparison for $8\times$ SR on \textsc{BSDS100} and \textsc{Urban100}.
		}
		\label{fig:urban100-x8}
		\vspace{-5mm}
	\end{figure*}
	
\begin{table*}[t]
	\centering
	\caption{
		Quantitative evaluation of state-of-the-art SR algorithms: average PSNR/SSIM/IFC for scale factors $2\times$, $4\times$ and $8\times$.
		\red{\textbf{Red}} text indicates the best and \blue{\underline{blue}} text indicates the second best performance.
	}
	\vspace{-0.2cm}
	\label{tab:quality} 
	\resizebox{\textwidth}{!}{
		\begin{tabular}{lcccccc}
			\toprule
			\multirow{2}{*}{Algorithm}
			&
			\multirow{2}{*}{Scale}
			& 
			\textsc{Set5} & \textsc{Set14} & \textsc{BSDS100} & \textsc{Urban100} & \textsc{manga109} \\
			& 
			&
			PSNR / SSIM / IFC & 
			PSNR / SSIM / IFC & 
			PSNR / SSIM / IFC & 
			PSNR / SSIM / IFC & 
			PSNR / SSIM / IFC \\
			\midrule
			Bicubic & 2 &
			33.65 / 0.930 / 6.166 &
			30.34 / 0.870 / 6.126 &
			29.56 / 0.844 / 5.695 &
			26.88 / 0.841 / 6.319 &
			30.84 / 0.935 / 6.214 \\
			A+~\cite{A+} & 2 &
			36.54 / 0.954 / \blue{\underline{8.715}} &
			32.40 / 0.906 / \blue{\underline{8.201}} &
			31.22 / 0.887 / \blue{\underline{7.464}} &
			29.23 / 0.894 / 8.440 &
			35.33 / 0.967 / 8.906 \\
			SRCNN~\cite{SRCNN} & 2 &
			36.65 / 0.954 / 8.165 &
			32.29 / 0.903 / 7.829 &
			31.36 / 0.888 / 7.242 &
			29.52 / 0.895 / 8.092 &
			35.72 / 0.968 / 8.471 \\
			FSRCNN~\cite{FSRCNN} & 2 &
			36.99 / 0.955 / 8.200 &
			32.73 / 0.909 / 7.843 &
			31.51 / 0.891 / 7.180 &
			29.87 / 0.901 / 8.131 &
			36.62 / 0.971 / 8.587 \\
			SelfExSR~\cite{Huang-CVPR-2015} & 2 &
			36.49 / 0.954 / 8.391 &
			32.44 / 0.906 / 8.014 &
			31.18 / 0.886 / 7.239 &
			29.54 / 0.897 / 8.414 &
			35.78 / 0.968 / 8.721 \\
			RFL~\cite{RFL} & 2 &
			36.55 / 0.954 / 8.006 &
			32.36 / 0.905 / 7.684 &
			31.16 / 0.885 / 6.930 &
			29.13 / 0.891 / 7.840 &
			35.08 / 0.966 / 8.921 \\
			SCN~\cite{SCN} & 2 &
			36.52 / 0.953 / 7.358 &
			32.42 / 0.904 / 7.085 &
			31.24 / 0.884 / 6.500 &
			29.50 / 0.896 / 7.324 &
			35.47 / 0.966 / 7.601 \\
			VDSR~\cite{VDSR} & 2 &
			\blue{\underline{37.53}} / 0.958 / 8.190 &
			32.97 / \red{\textbf{0.913}} / 7.878 &
			\red{\textbf{31.90}} / \red{\textbf{0.896}} / 7.169 &
			\red{\textbf{30.77}} / \red{\textbf{0.914}} / 8.270 &
			37.16 / \blue{\underline{0.974}} / 9.120 \\
			DRCN~\cite{DRCN} & 2 &
			\red{\textbf{37.63}} / \blue{\underline{0.959}} / 8.326 &
			\blue{\underline{32.98}} / 0.913 / 8.025 &
			\blue{\underline{31.85}} / 0.894 / 7.220 &
			\blue{\underline{30.76}} / \blue{\underline{0.913}} / 8.527 &
			\red{\textbf{37.57}} / 0.973 / \red{\textbf{9.541}} \\
			LapSRN (ours $2\times$) & 2 &
			37.52 / \red{\textbf{0.959}} / \red{\textbf{9.010}} &
			\red{\textbf{33.08}} / \blue{\underline{0.913}} / \red{\textbf{8.505}} &
			31.80 / \blue{\underline{0.895}} / \red{\textbf{7.715}} &
			30.41 / 0.910 / \red{\textbf{8.907}} &
			\blue{\underline{37.27}} / \red{\textbf{0.974}} / \blue{\underline{9.481}} \\
			LapSRN (ours $8\times$) & 2 &
			37.25 / 0.957 / 8.527 &
			32.96 / 0.910 / 8.140 &
			31.68 / 0.892 / 7.430 &
			30.25 / 0.907 / \blue{\underline{8.564}} &
			36.73 / 0.972 / 8.933 \\
			\midrule
			Bicubic & 4 &
			28.42 / 0.810 / 2.337 &
			26.10 / 0.704 / 2.246 &
			25.96 / 0.669 / 1.993 &
			23.15 / 0.659 / 2.386 &
			24.92 / 0.789 / 2.289 \\
			A+~\cite{A+} & 4 &
			30.30 / 0.859 / 3.260 &
			27.43 / 0.752 / 2.961 &
			26.82 / 0.710 / 2.564 &
			24.34 / 0.720 / 3.218 &
			27.02 / 0.850 / 3.177 \\
			SRCNN~\cite{SRCNN} & 4 &
			30.49 / 0.862 / 2.997 &
			27.61 / 0.754 / 2.767 &
			26.91 / 0.712 / 2.412 &
			24.53 / 0.724 / 2.992 &
			27.66 / 0.858 / 3.045 \\
			FSRCNN~\cite{FSRCNN} & 4 &
			30.71 / 0.865 / 2.994 &
			27.70 / 0.756 / 2.723 &
			26.97 / 0.714 / 2.370 &
			24.61 / 0.727 / 2.916 &
			27.89 / 0.859 / 2.950 \\
			SelfExSR~\cite{Huang-CVPR-2015} & 4 &
			30.33 / 0.861 / 3.249 &
			27.54 / 0.756 / 2.952 &
			26.84 / 0.712 / 2.512 &
			24.82 / 0.740 / 3.381 &
			27.82 / 0.865 / 3.358 \\
			RFL~\cite{RFL} & 4 &
			30.15 / 0.853 / 3.135 &
			27.33 / 0.748 / 2.853 &
			26.75 / 0.707 / 2.455 &
			24.20 / 0.711 / 3.000 &
			26.80 / 0.840 / 3.055 \\
			SCN~\cite{SCN} & 4 &
			30.39 / 0.862 / 2.911 &
			27.48 / 0.751 / 2.651 &
			26.87 / 0.710 / 2.309 &
			24.52 / 0.725 / 2.861 &
			27.39 / 0.856 / 2.889 \\
			VDSR~\cite{VDSR} & 4 &
			31.35 / 0.882 / 3.496 &
			28.03 / \blue{\underline{0.770}} / 3.071 &
			\blue{\underline{27.29}} / \blue{\underline{0.726}} / 2.627 &
			\blue{\underline{25.18}} / \blue{\underline{0.753}} / 3.405 &
			28.82 / 0.886 / 3.664 \\
			DRCN~\cite{DRCN} & 4 &
			\blue{\underline{31.53}} / \blue{\underline{0.884}} / \blue{\underline{3.502}} &
			28.04 / 0.770 / 3.066 &
			27.24 / 0.724 / 2.587 &
			25.14 / 0.752 / 3.412 &
			\blue{\underline{28.97}} / \blue{\underline{0.886}} / \blue{\underline{3.674}} \\
			LapSRN (ours $4\times$) & 4 &
			\red{\textbf{31.54}} / \red{\textbf{0.885}} / \red{\textbf{3.559}} &
			\red{\textbf{28.19}} / \red{\textbf{0.772}} / \red{\textbf{3.147}} &
			\red{\textbf{27.32}} / \red{\textbf{0.728}} / \red{\textbf{2.677}} &
			\red{\textbf{25.21}} / \red{\textbf{0.756}} / \red{\textbf{3.530}} &
			\red{\textbf{29.09}} / \red{\textbf{0.890}} / \red{\textbf{3.729}} \\
			LapSRN (ours $8\times$) & 4 &
			31.33 / 0.881 / 3.491 &
			\blue{\underline{28.06}} / 0.768 / \blue{\underline{3.100}} &
			27.22 / 0.724 / \blue{\underline{2.660}} &
			25.02 / 0.747 / \blue{\underline{3.426}} &
			28.68 / 0.882 / 3.595 \\
			\midrule
			Bicubic & 8 &
			24.39 / 0.657 / 0.836 &
			23.19 / 0.568 / 0.784 &
			23.67 / 0.547 / 0.646 &
			20.74 / 0.515 / 0.858 &
			21.47 / 0.649 / 0.810 \\
			A+~\cite{A+} & 8 &
			25.52 / 0.692 / 1.077 &
			23.98 / 0.597 / 0.983 &
			24.20 / 0.568 / 0.797 &
			21.37 / 0.545 / 1.092 &
			22.39 / 0.680 / 1.056 \\
			SRCNN~\cite{SRCNN} & 8 &
			25.33 / 0.689 / 0.938 &
			23.85 / 0.593 / 0.865 &
			24.13 / 0.565 / 0.705 &
			21.29 / 0.543 / 0.947 &
			22.37 / 0.682 / 0.940 \\
			FSRCNN~\cite{FSRCNN} & 8 &
			25.41 / 0.682 / 0.989 &
			23.93 / 0.592 / 0.928 &
			24.21 / 0.567 / 0.772 &
			21.32 / 0.537 / 0.986 &
			22.39 / 0.672 / 0.977 \\
			SelfExSR~\cite{Huang-CVPR-2015} & 8 &
			25.52 / 0.704 / \blue{\underline{1.131}} &
			24.02 / 0.603 / 1.001 &
			24.18 / 0.568 / 0.774 &
			\blue{\underline{21.81}} / \blue{\underline{0.576}} / \blue{\underline{1.283}} &
			\blue{\underline{22.99}} / \blue{\underline{0.718}} / \blue{\underline{1.244}} \\
			RFL~\cite{RFL} & 8 &
			25.36 / 0.677 / 0.985 &
			23.88 / 0.588 / 0.910 &
			24.13 / 0.562 / 0.741 &
			21.27 / 0.535 / 0.978 &
			22.27 / 0.668 / 0.968 \\
			SCN~\cite{SCN} & 8 &
			25.59 / 0.705 / 1.063 &
			24.11 / 0.605 / 0.967 &
			24.30 / 0.573 / 0.777 &
			21.52 / 0.559 / 1.074 &
			22.68 / 0.700 / 1.073 \\
			VDSR~\cite{VDSR} & 8 &
			\blue{\underline{25.72}} / \blue{\underline{0.711}} / 1.123 &
			\blue{\underline{24.21}} / \blue{\underline{0.609}} / \blue{\underline{1.016}} &
			\blue{\underline{24.37}} / \blue{\underline{0.576}} / \blue{\underline{0.816}} &
			21.54 / 0.560 / 1.119 &
			22.83 / 0.707 / 1.138 \\
			LapSRN (ours $8\times$) & 8 &
			\red{\textbf{26.14}} / \red{\textbf{0.738}} / \red{\textbf{1.302}} &
			\red{\textbf{24.44}} / \red{\textbf{0.623}} / \red{\textbf{1.134}} &
			\red{\textbf{24.54}} / \red{\textbf{0.586}} / \red{\textbf{0.893}} &
			\red{\textbf{21.81}} / \red{\textbf{0.581}} / \red{\textbf{1.288}} &
			\red{\textbf{23.39}} / \red{\textbf{0.735}} / \red{\textbf{1.352}} \\
			\bottomrule 
		\end{tabular}
	}
	\vspace{-4mm}
\end{table*}

	\subsection{Comparisons with the state-of-the-arts} 
	We compare the proposed LapSRN with 8 state-of-the-art SR algorithms: A+~\cite{A+}, SRCNN~\cite{SRCNN}, FSRCNN~\cite{FSRCNN}, SelfExSR~\cite{Huang-CVPR-2015}, RFL~\cite{RFL}, SCN~\cite{SCN}, VDSR~\cite{VDSR} and DRCN~\cite{DRCN}.
	We carry out extensive experiments using 5 datasets: \textsc{Set5}~\cite{Bevilacqua-BMVC-2012}, \textsc{Set14}~\cite{Zeyde-2010}, \textsc{BSDS100}~\cite{BSDS}, \textsc{Urban100}~\cite{Huang-CVPR-2015} and \textsc{manga109}~\cite{manga109}.
	Among these datasets, \textsc{Set5}, \textsc{Set14} and \textsc{BSDS100} consist of natural scenes; 
	\textsc{Urban100} contains challenging urban scenes images with details in different frequency bands;
	and \textsc{manga109} is a dataset of Japanese manga.
	We train the LapSRN until the learning rate decreases to $1e-6$ and the training time is around three days on a Titan X GPU.

	We evaluate the SR images with three commonly used image quality metrics: PSNR, SSIM~\cite{SSIM}, and IFC~\cite{IFC}.
	\tabref{quality} shows quantitative comparisons for $2\times$, $4\times$ and $8\times$ SR.
	Our LapSRN performs favorably against existing methods on most datasets.
	In particular, our algorithm achieves higher IFC values, which has been shown to be correlated well with human perception of image super-resolution~\cite{Yang-ECCV-2014}.
	%
	We note that the best results can be achieved by training with 
	specific scale factors (Ours $2\times$ and Ours $4\times$).
	As the intermediate convolutional layers are trained to minimize the prediction errors for both the corresponding level and higher levels, 
	the intermediate predictions of our $8\times$ model are slightly inferior to our $2\times$ and $4\times$ models.
	Nevertheless, our $8\times$ model provides a competitive performance to the state-of-the-art methods in $2\times$ and $4\times$ SR.

	In \figref{result-4x}, we show visual comparisons on \textsc{Urban100}, \textsc{BSDS100} and \textsc{Manga109} with the a scale factor of $4\times$.
	Our method accurately reconstructs parallel straight lines and grid patterns such as windows and the stripes on tigers.
	We observe that methods using the bicubic upsampling for pre-processing generate results with noticeable artifacts~\cite{SRCNN,VDSR,RFL,A+,SCN}.
	In contrast, our approach effectively suppresses such artifacts through progressive reconstruction and the robust loss function.

	For $8\times$ SR, we re-train the model of A+, SRCNN, FSRCNN, RFL and VDSR using the publicly available code\footnote{We do not re-train DRCN because the training code is not available.}.
	Both SelfExSR and SCN methods can handle different scale factors using progressive reconstruction.
	We show $8\times$ SR results on \textsc{BSDS100} and \textsc{Urban100} in~\figref{urban100-x8}.
	For $8\times$ SR, it is challenging to predict HR images from bicubic-upsampled images~\cite{SRCNN,VDSR,A+} or using one-step upsampling~\cite{FSRCNN}.
	The state-of-the-art methods do not super-resolve the fine structures well.
	In contrast, the LapSRN reconstructs high-quality HR images at a relatively fast speed.
	We present SR images generated by all the evaluated methods in the supplementary material.

	\subsection{Execution time}
	\label{sec:time}
\begin{figure}[t]
		\centering
		\includegraphics[width=0.86\columnwidth]{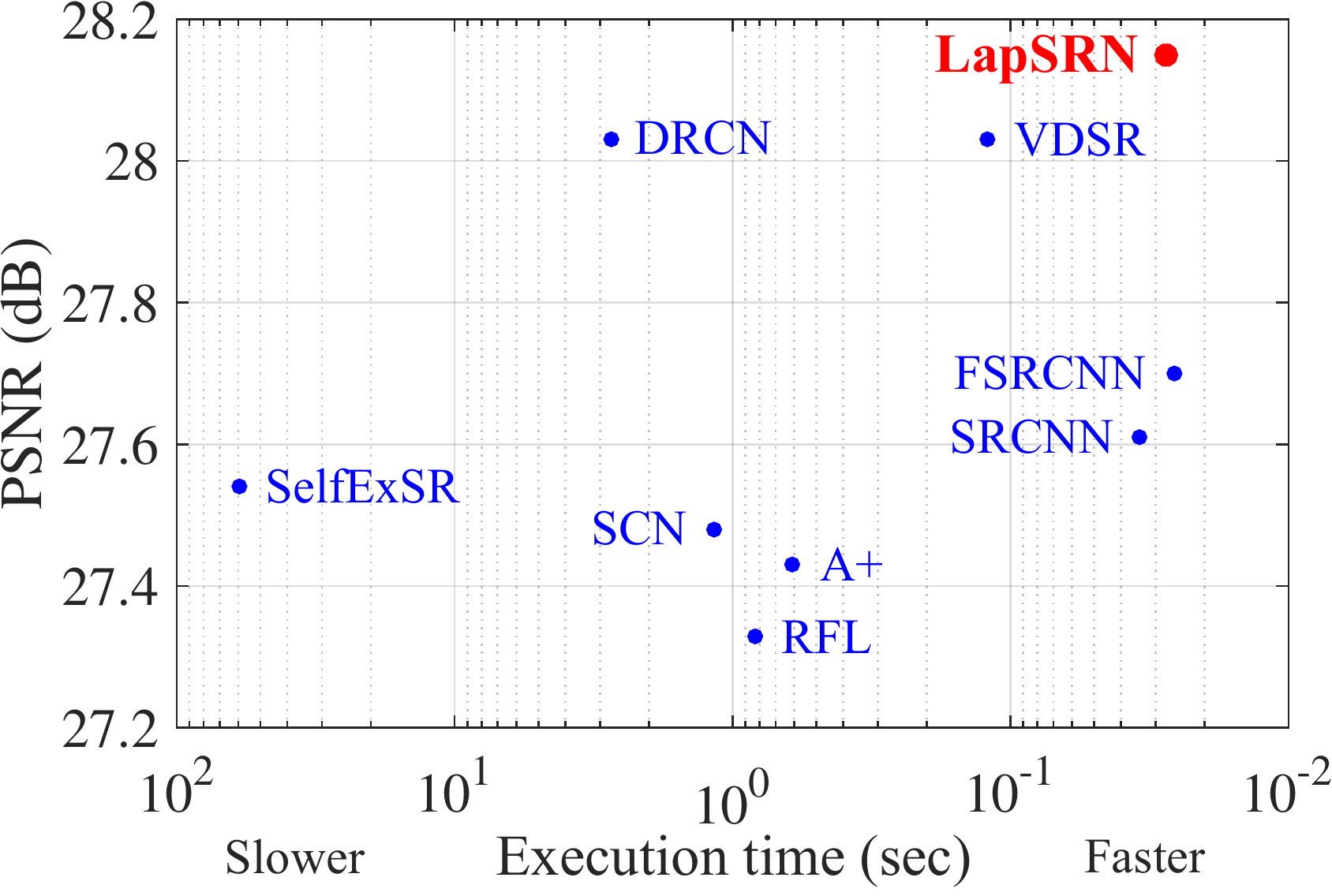}
		\vspace{-2mm}
		\caption{
			Speed and accuracy trade-off.
			The results are evaluated on \textsc{Set14} with the scale factor $4\times$. 
			The LapSRN generates SR images efficiently and accurately.
		}
		\label{fig:time-chart}
		\vspace{-4mm}
	\end{figure}

	We use the original codes of state-of-the-art methods to evaluate the runtime on the same machine with 3.4 GHz Intel i7 CPU (64G RAM) and NVIDIA Titan X GPU (12G Memory).
	Since the codes of SRCNN and FSRCNN for testing are based on CPU implementations, we reconstruct these models in MatConvNet with the same network weights to measure the run time on GPU.
	\figref{time-chart} shows the trade-offs between the run time and performance (in terms of PSNR) on \textsc{Set14} for $4\times$ SR.
	The speed of the proposed LapSRN is faster than all the existing methods except FSRCNN.
	We present detailed evaluations on run time of all evaluated datasets in the supplementary material.

	\begin{figure*}
		\footnotesize
		\centering
		\begin{tabular}{cccc}
			\hspace{-0.35cm}
			\begin{adjustbox}{valign=t}
				\begin{tabular}{c}
					\includegraphics[width=0.4\columnwidth]{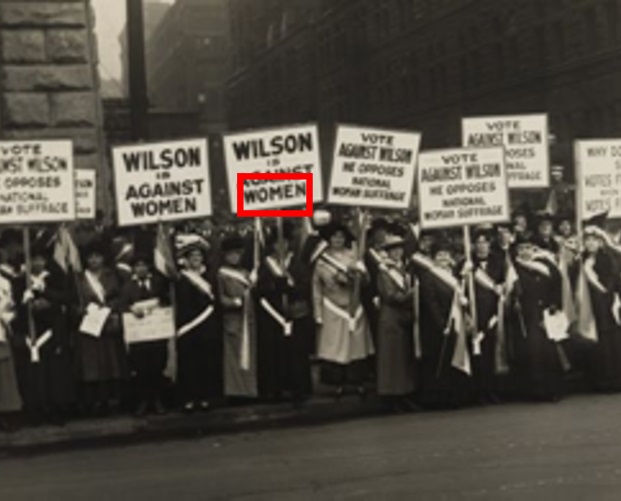}
					\\
					Ground-truth HR
				\end{tabular}
			\end{adjustbox}
			\hspace{-0.35cm}
			\begin{adjustbox}{valign=t}
				\begin{tabular}{cc}
					\includegraphics[width=0.265\columnwidth]{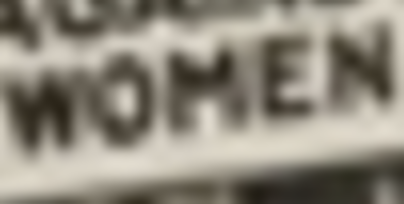} \hspace{-0.3cm} &
					\includegraphics[width=0.265\columnwidth]{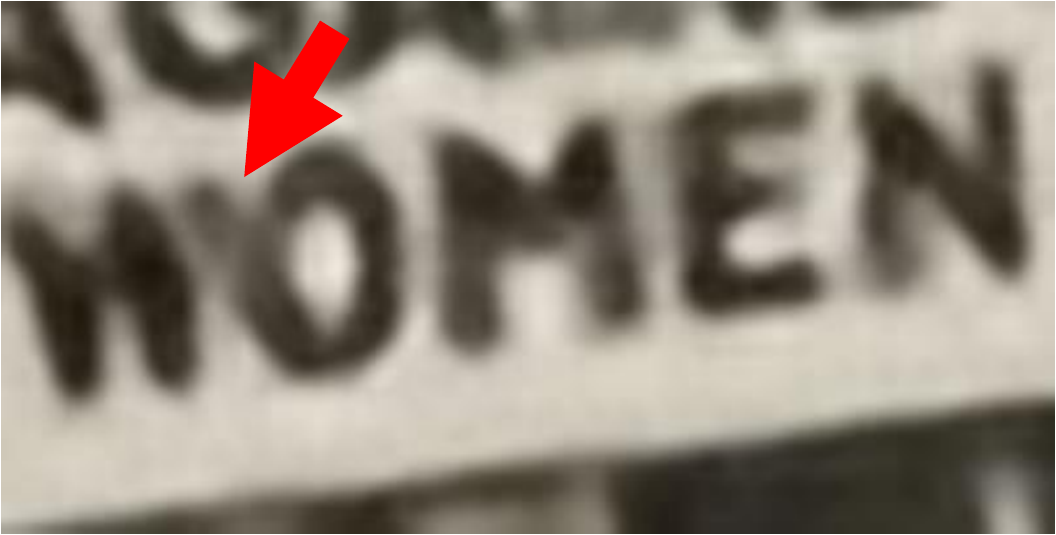}
					\\
					Bicubic \hspace{-0.3cm} &
					FSRCNN~\cite{FSRCNN}
					\\
					\includegraphics[width=0.265\columnwidth]{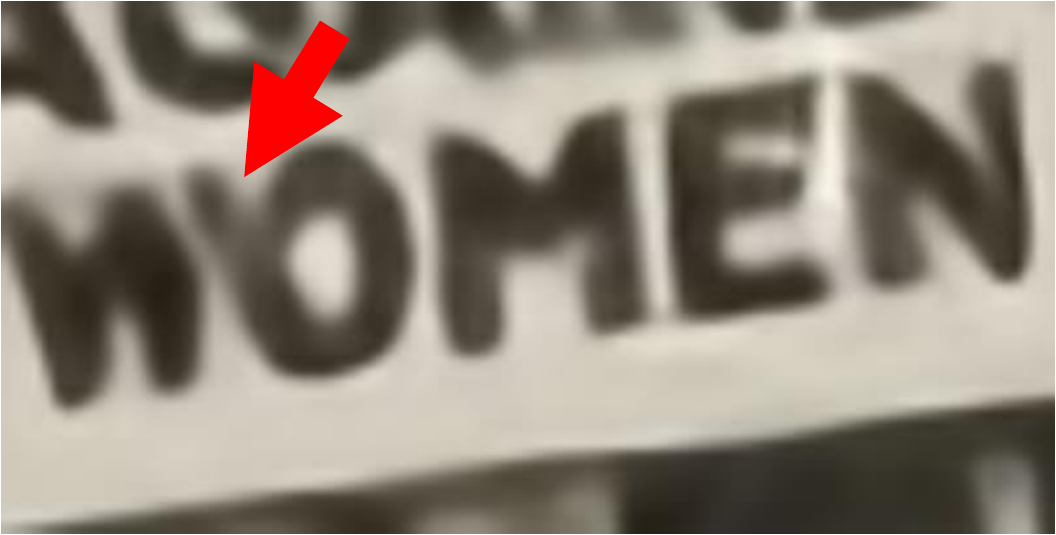} \hspace{-0.3cm} &
					\includegraphics[width=0.265\columnwidth]{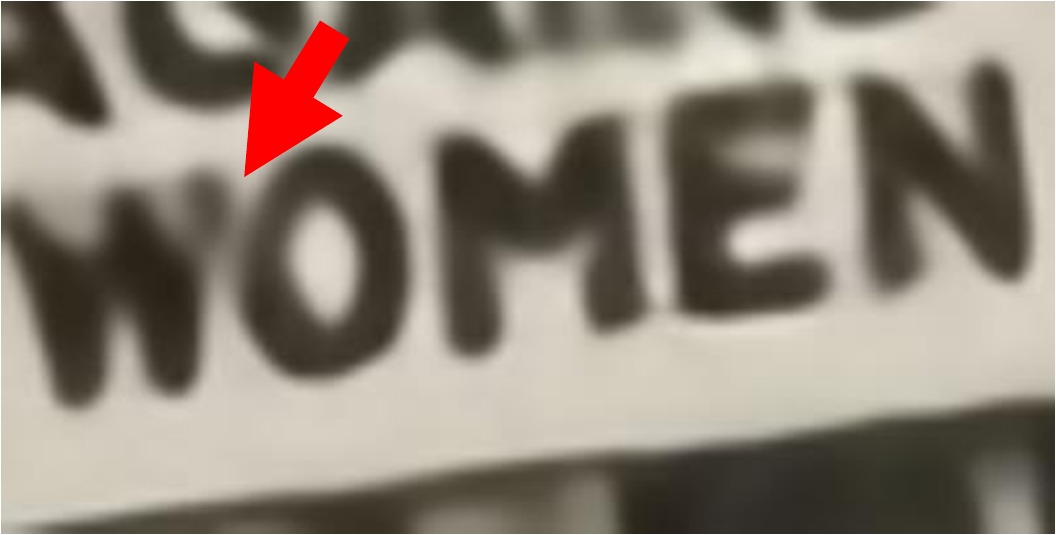} 
					\\
					\hspace{0.3cm}VDSR~\cite{VDSR}&
					LapSRN (ours)
				\end{tabular}
			\end{adjustbox}
			\hspace{0.35cm}
			\begin{adjustbox}{valign=t}
				\begin{tabular}{c}
					\includegraphics[width=0.4\columnwidth]{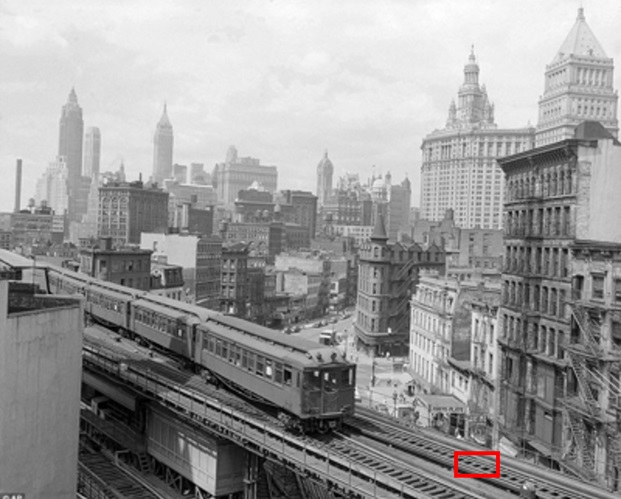}
					\\
					Ground-truth HR
				\end{tabular}
			\end{adjustbox}
			\hspace{-0.35cm}
			\begin{adjustbox}{valign=t}
				\begin{tabular}{cc}
					\includegraphics[width=0.265\columnwidth]{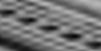} \hspace{-0.3cm} &
					\includegraphics[width=0.265\columnwidth]{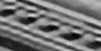}
					\\
					Bicubic \hspace{-0.3cm} &
					FSRCNN~\cite{FSRCNN}
					\\
					\includegraphics[width=0.265\columnwidth]{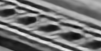} \hspace{-0.3cm} &
					\includegraphics[width=0.265\columnwidth]{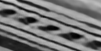} 
					\\
					\hspace{0.3cm}VDSR~\cite{VDSR}&
					LapSRN (ours)
				\end{tabular}
			\end{adjustbox}
		\end{tabular}
		\vspace{-2mm}
		\caption{
			Comparison of real-world photos for $4\times$ SR. 
			We note that the ground truth HR images and the blur kernels are not available in these cases.
			%
			On the left image, our method super-resolves the letter ``W'' accurately while VDSR incorrectly connects the stroke with the letter ``O''.
			On the right image, our method reconstructs the rails without the ringing artifacts.
		}
		\label{fig:historical}
		\vspace{-4mm}
	\end{figure*}

	\subsection{Super-resolving real-world photos}
	%
	We demonstrate an application of super-resolving historical photographs with JPEG compression artifacts.
	In these cases, neither the ground-truth images nor the downsampling kernels are available. 
	As shown in~\figref{historical}, our method can reconstruct sharper and more accurate images than the state-of-the-art approaches.
	%
	
	\subsection{Super-resolving video sequences}
\begin{figure}
		\footnotesize
		\centering
		\begin{tabular}{cc}
			\hspace{-0.32cm}
			\begin{adjustbox}{valign=t}
				\begin{tabular}{c}
					\includegraphics[width=0.4\columnwidth]{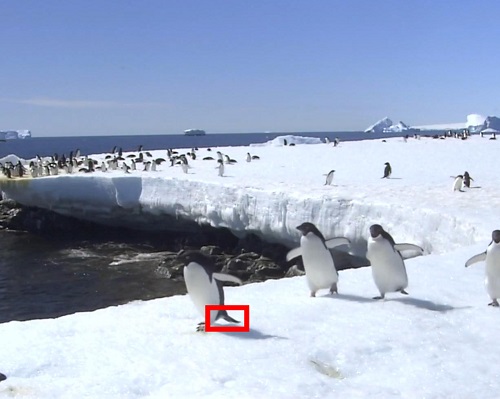}
					\\
					Ground-truth HR
				\end{tabular}
			\end{adjustbox}
			\hspace{-0.4cm}
			\begin{adjustbox}{valign=t}
				\begin{tabular}{cc}
					\includegraphics[width=0.265\columnwidth]{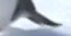} \hspace{-0.3cm} &
					\includegraphics[width=0.265\columnwidth]{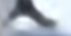}
					\\
					\hspace{0.2cm} HR &
					SRCNN~\cite{SRCNN}
					\\
					\includegraphics[width=0.265\columnwidth]{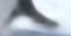} \hspace{-0.3cm} &
					\includegraphics[width=0.265\columnwidth]{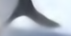} 
					\\
					\hspace{0.2cm} VDSR~\cite{VDSR}&
					LapSRN (ours)
				\end{tabular}
			\end{adjustbox}
		\end{tabular}
		\vspace{-2mm}
		\caption{
			Visual comparison on a video frame with a spatial resolution of $1200\times800$ for $8\times$ SR.
			Our method provides more clean and sharper results than existing methods.
		}
		\label{fig:video_8x}
		\vspace{-3mm}
	\end{figure}

	%
	%
	We conduct frame-based SR experiments on two video sequences from~\cite{Liao-ICCV-2015} with a spatial resolution of $1200 \times 800$ pixels.\footnote{Our method is \emph{not} a video super-resolution algorithm as temporal coherence or motion blur are not considered.}
	We downsample each frame by $8\times$, and then apply super-resolution frame by frame for $2\times$, $4\times$ and $8\times$, respectively. 
	The computational cost depends on the size of \textit{input} images since we extract features from the LR space.
	On the contrary, the speed of SRCNN and VDSR is limited by the size of \textit{output} images.
	Both FSRCNN and our approach achieve real-time performance (\ie, over 30 frames per second) on all upsampling scales.
	In contrast, the FPS is 8.43 for SRCNN and 1.98 for VDSR on $8\times$ SR.
	\figref{video_8x} visualizes results of $8\times$ SR on one representative frame. 
	%

	\subsection{Limitations}
\begin{figure}
		\footnotesize
		\centering
		\begin{tabular}{cc}
			\hspace{-0.35cm}
			\begin{adjustbox}{valign=t}
				\begin{tabular}{c}
					\includegraphics[width=0.4\columnwidth]{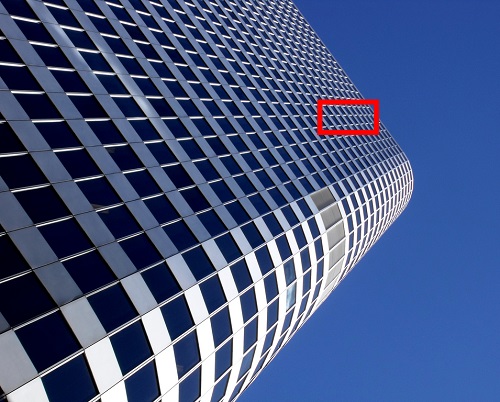}
					\\
					Ground-truth HR
				\end{tabular}
			\end{adjustbox}
			\hspace{-0.35cm}
			\begin{adjustbox}{valign=t}
				\begin{tabular}{cc}
					\includegraphics[width=0.265\columnwidth]{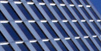} \hspace{-0.3cm} &
					\includegraphics[width=0.265\columnwidth]{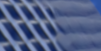}
					\\
					\hspace{0.2cm} HR &
					SelfExSR~\cite{Huang-CVPR-2015}
					\\
					\includegraphics[width=0.265\columnwidth]{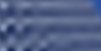} \hspace{-0.3cm} &
					\includegraphics[width=0.265\columnwidth]{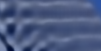} 
					\\
					\hspace{0.2cm} VDSR~\cite{VDSR}&
					LapSRN (ours)
				\end{tabular}
			\end{adjustbox}
		\end{tabular}
		\vspace{-2mm}
		\caption{A failure case for $8\times$ SR. Our method is not able to hallucinate details if the LR input image does not consist of sufficient amount of structure.}
		\label{fig:failure}
		\vspace{-3mm}
	\end{figure}

	While our model is capable of generating clean and sharp HR images on a large scale factor, e.g., $8\times$, it does not ``hallucinate'' fine details.
	As shown in~\figref{failure}, the top of the building is significantly blurred in the $8\times$ downscaled LR image.
	All SR algorithms fail to recover the fine structure except SelfExSR~\cite{Huang-CVPR-2015}, 
	which explicitly detects the 3D scene geometry and uses self-similarity to hallucinate the regular structure.
	This is a common limitation shared by parametric SR methods~\cite{SRCNN,FSRCNN,VDSR,DRCN}.
	%
	%
	Another limitation of the proposed network is the relative large model size.
	To reduce the number of parameters, one can replace the deep convolutional layers at each level with recursive layers.

	\section{Conclusions}
	\vspace{-0.1cm}
	In this work, we propose a deep convolutional network within a Laplacian pyramid framework for fast and accurate single-image super-resolution. 
	Our model progressively predicts high-frequency residuals in a coarse-to-fine manner. 
	By replacing the pre-defined bicubic interpolation with the learned transposed convolutional layers and optimizing the network with a robust loss function, the proposed LapSRN alleviates issues with undesired artifacts and reduces the computational complexity. 
	Extensive evaluations on benchmark datasets demonstrate that the proposed model performs favorably against the state-of-the-art SR algorithms in terms of visual quality and run time. 

	\section*{Acknowledgments}
	\vspace{-0.1cm}
	This work is supported in part by the NSF CAREER Grant $\#1149783$, gifts from Adobe and Nvidia. J.-B. Huang and N. Ahuja are supported in part by Office of Naval Research under Grant N00014-16-1-2314.
	
	\clearpage
	{\small
		\bibliographystyle{ieee}
		\bibliography{super_resolution}
	}
\end{document}